\documentclass[runningheads]{llncs}
\usepackage{graphicx}
\usepackage{subfigure}

\usepackage{tikz}
\usepackage{comment}
\usepackage{amsmath,amssymb} % define this before the line numbering.
\usepackage{color}

\usepackage[accsupp]{axessibility}  % Improves PDF readability for those with disabilities.

\usepackage{multirow}

\begin{document}
% \renewcommand\thelinenumber{\color[rgb]{0.2,0.5,0.8}\normalfont\sffamily\scriptsize\arabic{linenumber}\color[rgb]{0,0,0}}
% \renewcommand\makeLineNumber {\hss\thelinenumber\ \hspace{6mm} \rlap{\hskip\textwidth\ \hspace{6.5mm}\thelinenumber}}
% \linenumbers
\pagestyle{headings}
\mainmatter
\def\ECCVSubNumber{231}  % Insert your submission number here

\title{Normalized Feature Distillation for Semantic Segmentation} % Replace with your title

% INITIAL SUBMISSION 
\begin{comment}
\titlerunning{ECCV-22 submission ID \ECCVSubNumber} 
\authorrunning{ECCV-22 submission ID \ECCVSubNumber} 
\author{Anonymous ECCV submission}
\institute{Paper ID \ECCVSubNumber}
\end{comment}
%******************

% CAMERA READY SUBMISSION
% \begin{comment}
% \titlerunning{Abbreviated paper title}
% If the paper title is too long for the running head, you can set
% an abbreviated paper title here
%
\author{Tao Liu \and Xi Yang \and Chenshu Chen}
%
% \authorrunning{F. Author et al.}
% First names are abbreviated in the running head.
% If there are more than two authors, 'et al.' is used.
%
\institute{Hikvision Research Institute \\\email{\{liutao46,yangxi6,chenchenshu\}@hikvision.com}}

% \end{comment}
%******************
\maketitle

\begin{abstract}
As a promising approach in model compression, knowledge distillation improves the performance of a compact model by transferring the knowledge from a cumbersome one. The kind of knowledge used to guide the training of the student is important. Previous distillation methods in semantic segmentation strive to extract various forms of knowledge from the features, which involve elaborate manual design relying on prior information and have limited performance gains. In this paper, we propose a simple yet effective feature distillation method called normalized feature distillation (NFD), aiming to enable effective distillation with the original features without the need to manually design new forms of knowledge. The key idea is to prevent the student from focusing on imitating the magnitude of the teacher's feature response by normalization. Our method achieves state-of-the-art distillation results for semantic segmentation on Cityscapes, VOC 2012, and ADE20K datasets. Code will be available.

\keywords{Semantic Segmentation, Knowledge Distillation, Normalization}
\end{abstract}

\section{Introduction}

Recent works on backbone networks \cite{ResNet,DenseNet,ResNeSt,HRNet} and
segmentation frameworks \cite{PSPNet,DeepLab,OCRNet} have greatly improved
the performance of semantic segmentation. However, these high-performance models often require a lot of memory and computational overhead.
In real-time applications, lightweight models are preferred
due to limited resources. Designing lightweight networks \cite{MobileNet,ShuffleNet,ICNet,BiSeNet} and performing model compression \cite{XNOR-Net,Deep-Compression,KD} are feasible ways to address this problem. Knowledge distillation (KD) is a promising approach in model compression by distilling the knowledge of a cumbersome model (teacher) into a compact one (student).

Since features in the intermediate layers of deep neural networks contain rich representational information, feature distillation is a common approach in KD. FitNets \cite{FitNets} extends the idea of \cite{KD} to intermediate representations to improve the training of the student. The key idea is enabling the student to directly imitate the teacher's middle layer features. Inspired by this, a line of methods have been proposed to align features between the teacher and student in an indirect manner \cite{Survey}. They strive to extract various forms of manually designed knowledge from the features, such as attention map \cite{AT}, Gramian matrix \cite{FSP}, and pair-wise similarity \cite{SKD}. These methods typically rely on prior knowledge as in traditional feature engineering and suffer from information missing when performing transformations on original features, e.g., the channel information of features is lost in \cite{AT}. Although the performance improvement of FitNets \cite{FitNets} is limited, its advantage is that there is no information missing \cite{overhaul}.

Why does distillation with the original features not work as well as expected? \textit{We believe that forcing the student to imitate the exact same feature responses as the teacher is inefficient and unnecessary for distillation, but instead increases the learning difficulty}. Therefore, we propose a simple yet effective feature distillation method for semantic segmentation called normalized feature distillation (NFD), aiming to remove the magnitude of feature response by normalization during the distillation. Different from previous methods, we expect the student to capture valuable knowledge from the teacher's original features in an effective way. So our method doesn't involve elaborating various forms of knowledge.

Our main contributions are summarized as follows:
\begin{itemize}
    \item[$\bullet$] We propose a simple yet effective feature distillation method for semantic segmentation without the need to carefully design various forms of knowledge.
    \item[$\bullet$] We empirically show that simply encouraging the student to learn the normalized feature distribution from the teacher can yield superior distillation performance.
    \item[$\bullet$] We evaluate the proposed method on Cityscapes, VOC 2012 and ADE20K datasets. Extensive experiments demonstrate that our method significantly improves the performance of the baseline student model, achieving the state-of-the-art distillation results.
\end{itemize}

\section{Related Work}

\subsection{Knowledge Distillation for Semantic Segmentation}
Knowledge distillation has been extensively studied in the field of image classification, which has been proven to be a promising way to improve the performance of lightweight models. Subsequent researches tend to extend knowledge distillation to dense prediction tasks, which turn out to be more challenging, such as semantic segmentation \cite{Fast-Segmentation,Knowledge-Adaptation,SKD,IFVD,ICC,CWD}. Applying distillation methods in image classification to dense prediction tasks in a straightforward way may not yield satisfactory results. As a result, distillation methods for semantic segmentation have been proposed.

Xie \textit{et al.} \cite{Fast-Segmentation} use the Euclidean distance
between a pixel and its 8-neighbours to construct the so-called first-order knowledge to ensure that the segmented boundary information obtained by student and teacher are closed with each other. Liu \textit{et al.} \cite{SKD} distill the long-range dependency by computing the pair-wise similarity on the feature map and enforce high-order consistency between the outputs of the teacher and student through adversarial learning. Wang \textit{et al.} \cite{IFVD} propose to transfer the intra-class feature variation of the teacher to the student. Shu \textit{et al.} \cite{CWD} focus on channel information by softly aligning the activation of each channel between the teacher and student, which turns out to be more effective on logits than on features.

Most of previous methods either focus on extracting various forms of manually designed knowledge from features or mainly focus on logits distillation. Different from previous methods, we aim to perform effective distillation with original features, rather than designing new forms of knowledge.

\subsection{Normalization}
It is a fact that the input distribution of each layer changes during training neural networks, which slows down the training and makes it hard to train deep neural networks. Ioffe and Szegedy \cite{BN} refer to this phenomenon as internal covariate shift, and propose Batch Normalization (BN) to address the problem by normalizing the features by the mean and variance computed within a mini-batch. It has been shown to have the ability to facilitate optimization, and to enable convergence of very deep networks. However, the concept of “batch” is not always present, or it may change from time to time \cite{GN}. After that, some normalization methods without relying on the batch dimension have been proposed. Layer Normalization (LN) \cite{LN} operates along the channel dimension, which works well for the recurrent neural networks. Instance normalization (IN) \cite{IN}, typically used in image stylization, works in a similar way to BN but only for each sample. Group Normalization (GN) \cite{GN} divides the channels into groups and computes within each group the mean and variance for normalization, which works consistently over a wide range of batch sizes.

In this paper, we aim to remove the magnitude information of the original features by normalization along different dimensions as in \cite{BN,LN,IN}. But unlike them, the normalization in our method has no trainable scaling and shifting parameters.

\section{Method}

\subsection{Preliminary}
As it is simple and effective, the conventional KD loss on predictions proposed in \cite{KD} is a commonly used objective for many vision tasks. Treating semantic segmentation as a per-pixel classification problem, previous methods \cite{SKD,IFVD,CWD} adopt this loss to make the student imitate the teacher's per-pixel class probability, which can be formulated as:
\begin{align}
  \mathcal{L}_{kd}=D_\mathrm{KL}(\boldsymbol{Q}^t, \boldsymbol{Q}^s)\,,
\end{align}
where $\boldsymbol{Q}^t$ and $\boldsymbol{Q}^s$ denote the segmentation probability maps produced by the teacher and student, respectively, $D_\mathrm{KL}(\cdot)$ is the Kullback-Leibler divergence.

Let $\boldsymbol{F}^t \in \mathbb{R}^{B \times C \times H \times W}$ denotes the features of the teacher, where $B$ is the batch size, $C$ is the number of channels, $H$ and $W$ are the height and width. Similarly, $\boldsymbol{F}^s \in \mathbb{R}^{B \times C^\prime \times H^\prime \times W^\prime}$ denotes the features of the student. The general form of feature distillation loss can be formulated as:
\begin{align}
  \mathcal{L}_{feat}=D(T^t(\boldsymbol{F}^t), T^s(\boldsymbol{F}^s))\,,
  \label{eq:feat_kd}
\end{align}
where $T^t(\cdot)$ and $T^s(\cdot)$ are transformations applied to the teacher and student features, respectively, and $D(\cdot)$ measures the distance between features, which is usually the $L_1$ or $L_2$ distance. Since it is a common practice to make $\boldsymbol{F}^s$ has the same dimensions as $\boldsymbol{F}^t$ by convolution or upsampling, the dimension alignment will be done before $T^t(\cdot)$ and $T^s(\cdot)$ if necessary. 

Different feature distillation methods usually take different $T^t(\cdot)$ and $T^s(\cdot)$. For example, $T^t(\cdot)$ and $T^s(\cdot)$ can be summation of the features along the channel dimension in \cite{AT}.

\subsection{Normalized Feature Distillation}
The naive feature distillation proposed in \cite{FitNets} encourages the student to directly imitate the original features of the teacher. This means that $T^t(\cdot)$ and $T^s(\cdot)$ in Eq.~(\ref{eq:feat_kd}) are not needed, only dimension alignment for student features is required. Although there is no information missing in this way, its performance improvement is limited. We perform feature distillation in this naive way and analyze the feature differences between the teacher and student. 

\begin{figure}
    \centering
    \subfigure[]{
        \includegraphics[width=0.45\linewidth]{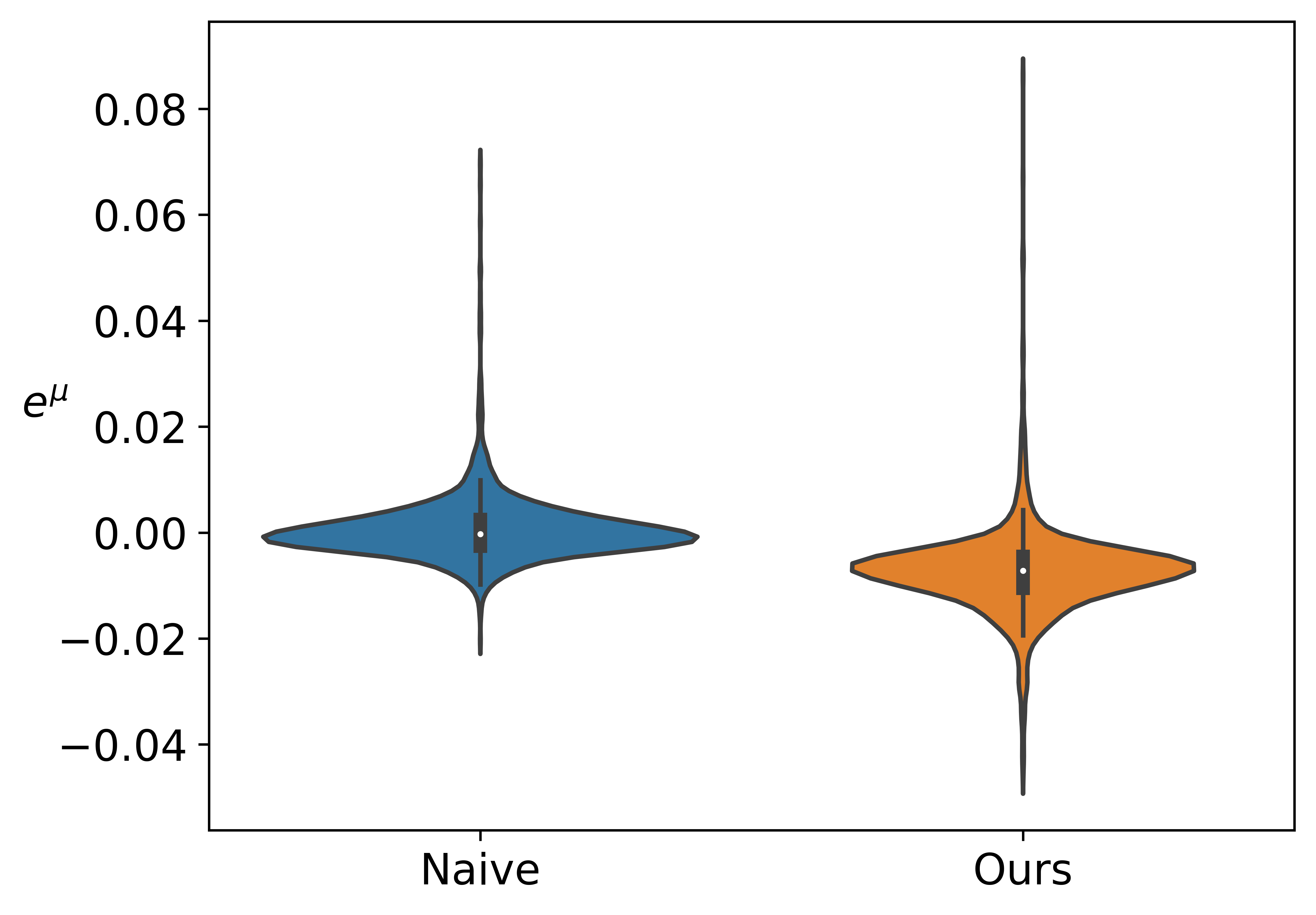}
    }
    \subfigure[]{
        \includegraphics[width=0.45\linewidth]{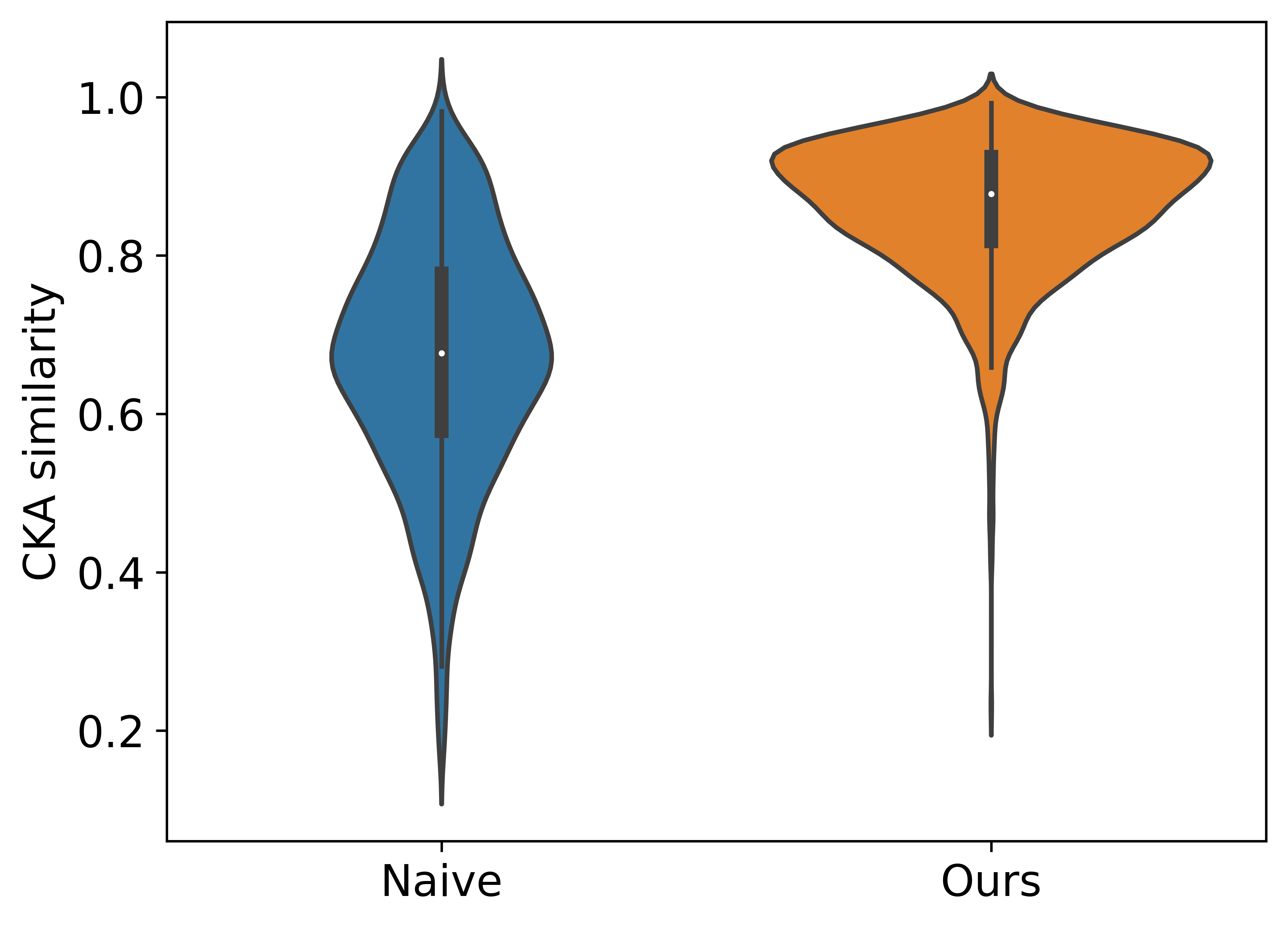}
    }
\caption{The feature differences between the teacher and student for naive feature distillation and our method on Cityscapes dataset. We analyze each channel's (a) error of the mean in Eq.~(\ref{eq:error_of_the_mean}) and (b) CKA similarity between the features of the teacher and student. The teacher model is PSPNet-R101, and the student model is PSPNet-R18. The features from the last layer of the backbone are used for distillation and to calculate the CKA similarity}
\label{fig:feat_mean_var_error}
\end{figure}

First, we compute the feature's mean for each channel, obtaining $\mu^t \in \mathbb{R}^C$ and $\mu^s \in \mathbb{R}^C$ for the teacher and student, respectively, and compute the following error:
\begin{align}
  e^{\mu}_i=\mu^t_i - \mu^s_i\,,
  \label{eq:error_of_the_mean}
\end{align}
where $i=1,2,...,C$. The distribution of $e^{\mu}$ is shown in the left part of Fig.~\ref{fig:feat_mean_var_error}(a). We can see the $e^{\mu}$ of most channels is close to zero, which means that the naive feature distillation forces the magnitude of feature response between the student and teacher to be close. 

In addition, we employ CKA \cite{CKA}, which is a similarity index enabling quantitative comparison of representations within and between networks, to measure the feature similarity between the teacher and student. CKA is invariant to orthogonal transformation and isotropic scaling of representations, so the magnitude of feature response does not affect the CKA similarity. We analyze the CKA similarity between the student and teacher features for each channel, as shown in the left part of Fig.~\ref{fig:feat_mean_var_error}(b). It is clear to see that the naive feature distillation does not result in a high similarity between the student and teacher features.

The above experiments demonstrate that distillation with original features in a naive way can result in the student focusing on learning the magnitude of feature response, failing to achieve a high feature similarity with the teacher. Building on this observation, we expect to ignore the magnitude information of the teacher's feature response during distillation and allow the student to learn effectively from the teacher's original features. Therefore, we propose a simple yet efficient feature distillation method called normalized feature distillation (NFD). First, we normalize the features of the teacher and student to remove their magnitude information, by making them have zero mean and unit variance. Then, we minimize the $L_2$ distance between the normalized features of the teacher and student, which can be formulated like in Eq.~(\ref{eq:feat_kd}) as:
\begin{align}
  \mathcal{L}_{nfd}=D(Norm(\boldsymbol{F}_t), Norm(\boldsymbol{F}_s))\,,
  \label{eq:nfd}
\end{align}
The $Norm$ in Eq.~(\ref{eq:nfd}) denotes normalization:
\begin{align}
  \boldsymbol{\hat{F}} = \frac{1}{\sigma}(\boldsymbol{F} - \mu)\,,
  \label{eq:norm}
\end{align}
where $ \boldsymbol{F} $ is the original feature, $\boldsymbol{\hat{F}}$ is the normalized feature, $\mu$ and $\sigma$ are feature's mean and standard deviation. We apply dimension alignment to student features before normalization. Inspired by different normalization methods, we can normalize the features along different dimensions, such as $(B,H,W)$ like BN \cite{BN}, $(C,H,W)$ like LN \cite{LN}, and $(H,W)$ like IN \cite{IN}. Unlike \cite{BN,LN,IN}, the normalization in Eq.~(\ref{eq:norm}) has no trainable scaling and shifting parameters. We will compare and analyze the performance of normalization along different dimensions in Section 4.4. We normalize the features along $(H,W)$ dimensions by default. 

Comparing with the naive feature distillation, the proposed method focuses on achieving high similarity with the teacher rather than imitating the magnitude of feature response, as shown in the right part of Fig.~\ref{fig:feat_mean_var_error}(a) and Fig.~\ref{fig:feat_mean_var_error}(b).

\begin{figure}
    \centering
    \includegraphics[width=0.9\linewidth]{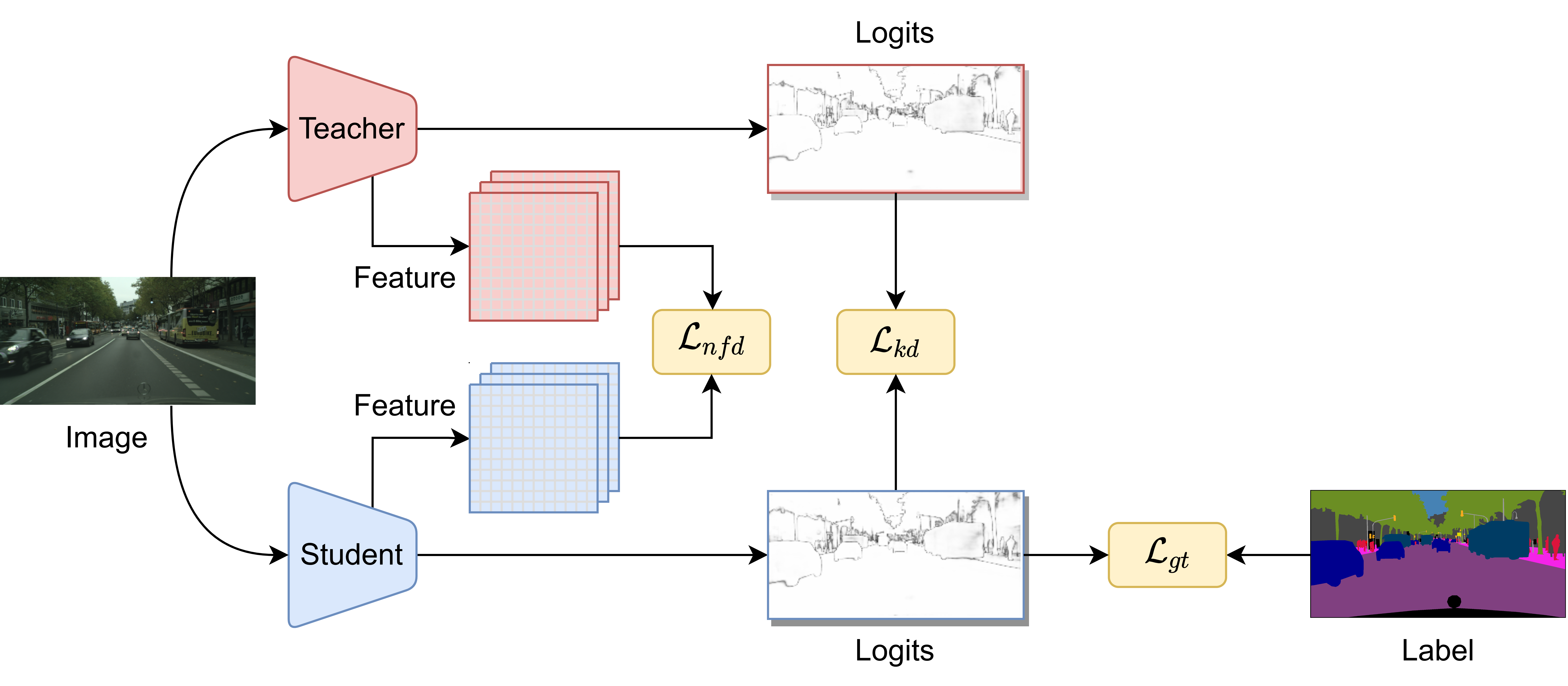}
\caption{The pipeline of applying our method to semantic segmentation. $\mathcal{L}_{nfd}$ is the proposed normalized feature distillation loss imposed on the intermediate layer features. $\mathcal{L}_{kd}$ is the conventional KD loss on logits. $\mathcal{L}_{gt}$ is the cross-entropy loss for semantic segmentation}
\label{fig:pipeline}
\end{figure}

The pipeline of applying our method to semantic segmentation is shown in Fig.~\ref{fig:pipeline}. Following the previous methods \cite{SKD,IFVD,CWD}, we adopt the conventional KD loss \cite{KD} on logits as well. Therefore, the total loss of our method can be formulated as:
\begin{align}
  \mathcal{L}= \mathcal{L}_{gt} + \lambda_1 \mathcal{L}_{kd} + \lambda_2 \mathcal{L}_{nfd} \,,
  \label{eq:total_loss}
\end{align}
where $\mathcal{L}_{gt}$ is the cross-entropy loss for semantic segmentation, $\lambda_1$ is set to 10 following \cite{SKD,IFVD,CWD}, $\lambda_2$ is set to 0.7.

\section{Experiments}
\subsection{Datasets}
\subsubsection{Cityscapes.} The Cityscapes \cite{Cityscapes} is a large-scale dataset for semantic urban scene understanding, with high quality pixel-level annotations of 5000 images in addition to a larger set of 19998 weakly annotated images. It contains 30 classes, and 19 of them are used for evaluation. The finely annotated 5000 images are divided into 2975, 500 and 1525 images for training, validation and testing. We only use the finely annotated dataset in our experiments.
\subsubsection{PASCAL VOC 2012.} The PASCAL VOC 2012 \cite{VOC} dataset contains 20 common objects and one background class with annotations on daily captured photos. We use the augmented dataset with extra coarse annotations provided by \cite{SBD} resulting in 10582, 1449 and 1456 images for training, validation and testing.
\subsubsection{ADE20K.} The ADE20K \cite{ADE20K} is a densely annotated dataset with the instances of stuff, objects, and parts, covering a diverse set of visual concepts in scenes. It contains 150 classes and is divided into 20210, 2000 and 3000 images for training, validation and testing. It is challenging due to its large number of classes and existence of multiple small objects in complex scenes. 

\subsection{Implementation Details}
\subsubsection{Network Architectures.} We adopt PSPNet \cite{PSPNet} with ResNet101 \cite{ResNet} backbone as the teacher model for all experiments. We adopt different segmentation models (PSPNet \cite{PSPNet} and DeepLabV3 \cite{DeepLabV3}) and backbones (ResNet18 \cite{ResNet} and MobileNetV2 \cite{MobileNetV2}) for the student model to verify the effectiveness of our method. 
\subsubsection{Training Details.} We use the pretrained teacher model and keep its parameters fixed during distillation. For the training of the student, we use Stochastic Gradient Descent (SGD) as the optimizer with a batch size of 16, a weight decay of 0.0005 and a momentum of 0.9. We use the “poly” learning rate policy where the learning rate equals to $base\_lr*(1 - \frac{iter}{max\_iter})^{power}$. We set the base learning rate to 0.01 and power to 0.9. We train 80k iterations for Cityscapes and VOC 2012 dataset and 160k iterations for ADE20K dataset. We apply random horizontal flipping, random scaling (from 0.5 to 2.0) and random cropping on the input images as data augmentation during training. The crop size for Cityscapes, VOC 2012 and ADE20K are $512 \times 1024$, $512\times512$ and $512\times512$, respectively. We use single scale testing for all datasets. Unless otherwise stated, the features from the last layer of the backbone are used for feature distillation in our method.

\setlength{\tabcolsep}{4pt}
\begin{table}[t]
    \begin{center}
    \caption{Comparison with the state-of-the-art methods on validation sets of Cityscapes, VOC 2012 and ADE20K. ``T'' denotes the teacher model, and ``S'' denotes the student model. ``R101'', ``R18'' and ``MV2'' denote ResNet101, ResNet18 and MobileNetV2, respectively. ``Baseline'' means training without distillation}
    \label{tab:sota_compare}
    \begin{tabular}{l|l|c|c|c}
        \hline
        \multirow{2}{*}{Model} & \multirow{2}{*}{Method} & \multicolumn{3}{c}{mIoU (\%)}\\
        \cline{3-5}
        & & Cityscapes & VOC 2012 & ADE20K\\
        \hline
        T: PSPNet-R101 & -- & 79.76 & 78.52 & 44.39\\
        \hline
        \multirow{5}{*}{S: PSPNet-R18} & Baseline & 72.65 & 71.35 & 35.03\\
        & SKD \cite{SKD} & 74.23 & 72.01 & 35.26\\
        & IFVD  \cite{IFVD} & 74.55 & 72.00 & 35.92\\
        & CWD \cite{CWD} & 75.91 & 73.07 & 36.78\\
        & NFD (Ours) & \textbf{76.63} & \textbf{75.38} & \textbf{39.09}\\
        \hline
        \multirow{5}{*}{S: PSPNet-MV2} & Baseline & 72.73 & 69.14 & 33.33\\
        & SKD \cite{SKD} & 72.90 & 69.62 & 33.39\\
        & IFVD \cite{IFVD} & 73.74 & 69.45 & 33.85\\
        & CWD \cite{CWD} & 74.73 & 71.28 & 35.26\\
        & NFD (Ours) & \textbf{75.50} & \textbf{73.43} & \textbf{37.01}\\
        \hline
        \multirow{5}{*}{S: DeepLabV3-R18} & Baseline & 74.96 & 71.98 & 37.19\\
        & SKD \cite{SKD} & 75.32 & 73.03 & 36.91\\
        & IFVD \cite{IFVD} & 76.01 & 72.87 & 37.66\\
        & CWD \cite{CWD} & 77.13 & 73.78 & 38.64\\
        & NFD (Ours) & \textbf{77.88} & \textbf{75.59} & \textbf{40.37}\\
        \hline
        \multirow{5}{*}{S: DeepLabV3-MV2} & Baseline & 73.98 & 69.92 & 35.14\\
        & SKD \cite{SKD} & 75.78 & 70.13 & 35.11\\
        & IFVD \cite{IFVD} & 75.24 & 70.32 & 35.35\\
        & CWD \cite{CWD} & 76.59 & 71.68 & 36.49\\
        & NFD (Ours) & \textbf{76.85} & \textbf{73.83} & \textbf{38.94}\\
        \hline
    \end{tabular}
    \end{center}
\end{table}
\setlength{\tabcolsep}{1.4pt}

\subsection{Comparison with the State-of-the-Art Methods}
We evaluate the proposed NFD and compare it with recent knowledge distillation methods in semantic segmentation on Cityscapes \cite{Cityscapes}, PASCAL VOC 2012 \cite{VOC} and ADE20K \cite{ADE20K} datasets. We re-implemented SKD \cite{SKD}, IFVD \cite{IFVD} and CWD \cite{CWD} based on their released code. The hyper-parameters related to distillation loss are set according to their recommended values. For fair comparison, all methods including NFD, SKD \cite{SKD}, IFVD \cite{IFVD} and CWD \cite{CWD} use exactly the same training and testing strategies as described in Section 4.2.

Table~\ref{tab:sota_compare} shows the results on various student models with different backbones (ResNet18 \cite{ResNet} and MobileNetV2 \cite{MobileNetV2}) and decoders (PPM \cite{PSPNet} and ASPP \cite{DeepLabV3}). NFD significantly improves the performance of baseline student models. For example, the performance gains for PSPNet-R18 brought by NFD are 3.98\%, 4.03\% and 4.06\% on Cityscapes, VOC 2012 and ADE20K, respectively. Although NFD utilizes the features from the last layer of the backbone network for distillation by default, the performance gains are not much affected by the backbone architecture. Specifically, NFD improves the performance of PSPNet-MV2 by 2.77\%, 4.29\% and 3.68\% on Cityscapes, VOC 2012 and ADE20K, respectively. In addition, NFD further narrows the performance gap between the teacher model and DeepLabV3-R18 which acts as a strong baseline student model.

More importantly, the proposed NFD consistently outperforms other methods by a large margin under various experimental setups, especially on VOC 2012 and ADE20K datasets. For example, NFD outperforms CWD \cite{CWD}, which was the previous state-of-the-art distillation method in semantic segmentation, by 2.31\%, 1.75\%, 1.73\% and 2.45\% when using PSPNet-R18, PSPNet-MV2, DeepLabV3-R18 and DeepLabV3-MV2 as student on more challenging ADE20K dataset. 

\subsection{Ablation Study}
In this section, we give extensive experiments to verify the effectiveness of our method and discuss the choice of some hyper-parameters. Ablation experiments are mainly conducted on Cityscapes and ADE20K datasets, with PSPNet-R101 as the teacher model and PSPNet-R18 as the student model.

\subsubsection{Feature Normalization.} As mentioned before, we can normalize the features along different dimensions when performing NFD, such as $(B,H,W)$ like BN \cite{BN}, $(C,H,W)$ like LN \cite{LN}, and $(H,W)$ like IN \cite{IN}. We conduct experiments to investigate the effects of different normalization dimensions. As shown in Table~\ref{tab:ablation_on_norm_dim}, the proposed NFD achieves similar performance under different normalization dimensions, and all normalization settings outperform the case without normalization. NFD without normalization is equivalent to the naive feature distillation in \cite{FitNets}. 

\setlength{\tabcolsep}{4pt}
\begin{table}
    \begin{center}
    \caption{Ablation study about the normalization dimensions of NFD. ``T'' denotes the teacher model, and ``S'' denotes the student model. ``w/o norm'' means NFD without normalization. The performance of the teacher and the baseline student is shown in the upper part of the table}
    \label{tab:ablation_on_norm_dim}
    \begin{tabular}{l|c|c|c}
        \hline
        \multirow{2}{*}{Model} & \multirow{2}{*}{Norm dimensions} & \multicolumn{2}{c}{mIoU (\%)}\\
        \cline{3-4}
        & & Cityscapes & ADE20K\\
        \hline
        T: PSPNet-R101 & -- & 79.76 & 44.39\\
        S: PSPNet-R18 & -- & 72.65 & 35.03\\
        \hline
        \multirow{5}{*}{S: PSPNet-R18} & w/o norm & 74.77 & 35.48\\
        & $(C,H,W)$ & 76.13 & \textbf{39.11}\\
        & $(B,H,W)$ & 76.26 & 38.83\\
        & $(H,W)$ & \textbf{76.63} & 39.09\\
        \hline
    \end{tabular}
    \end{center}
\end{table}
\setlength{\tabcolsep}{1.4pt}

\begin{figure}
    \centering
    \subfigure[w/o normalization]{
        \includegraphics[width=0.45\linewidth]{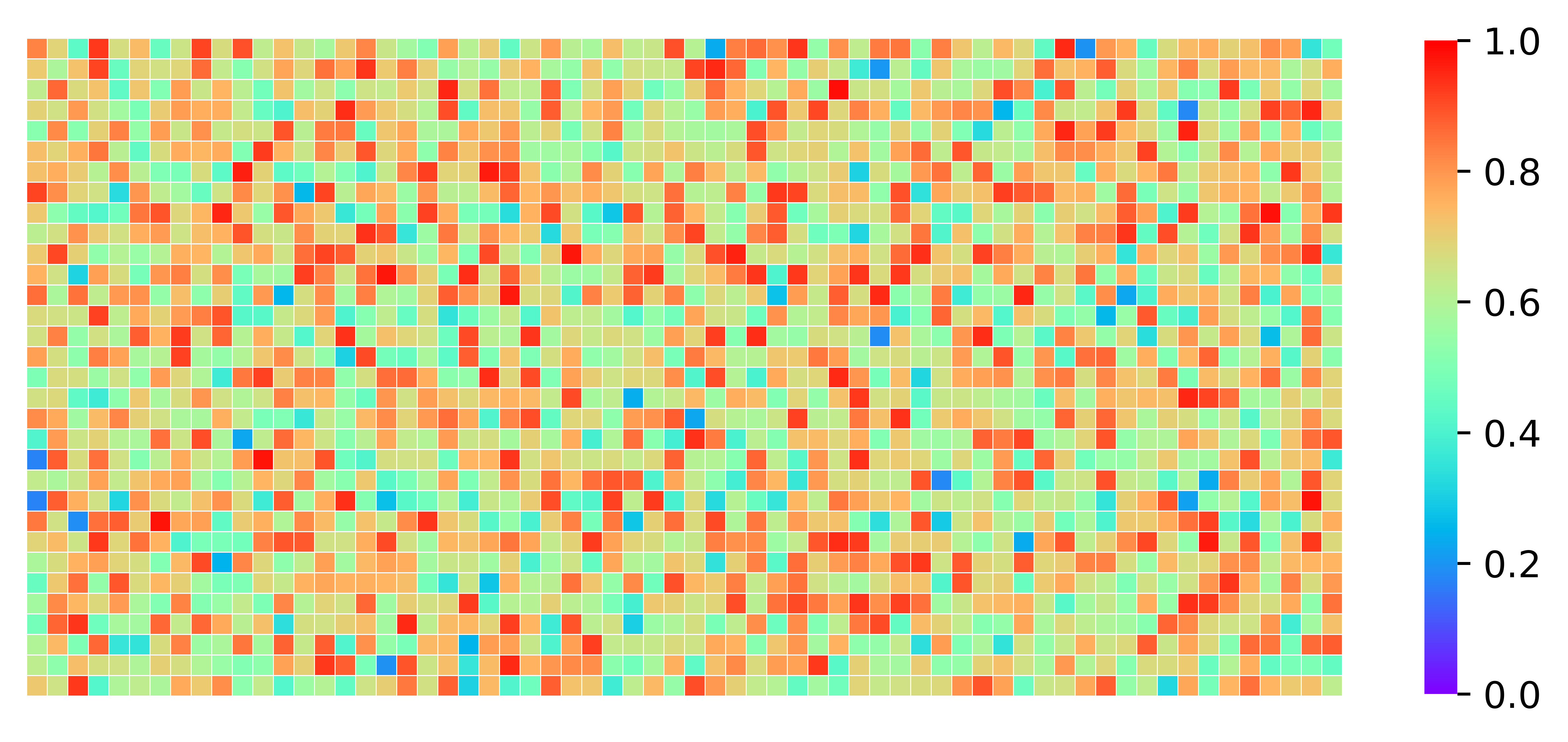}
    }
    \subfigure[$(C,H,W)$]{
        \includegraphics[width=0.45\linewidth]{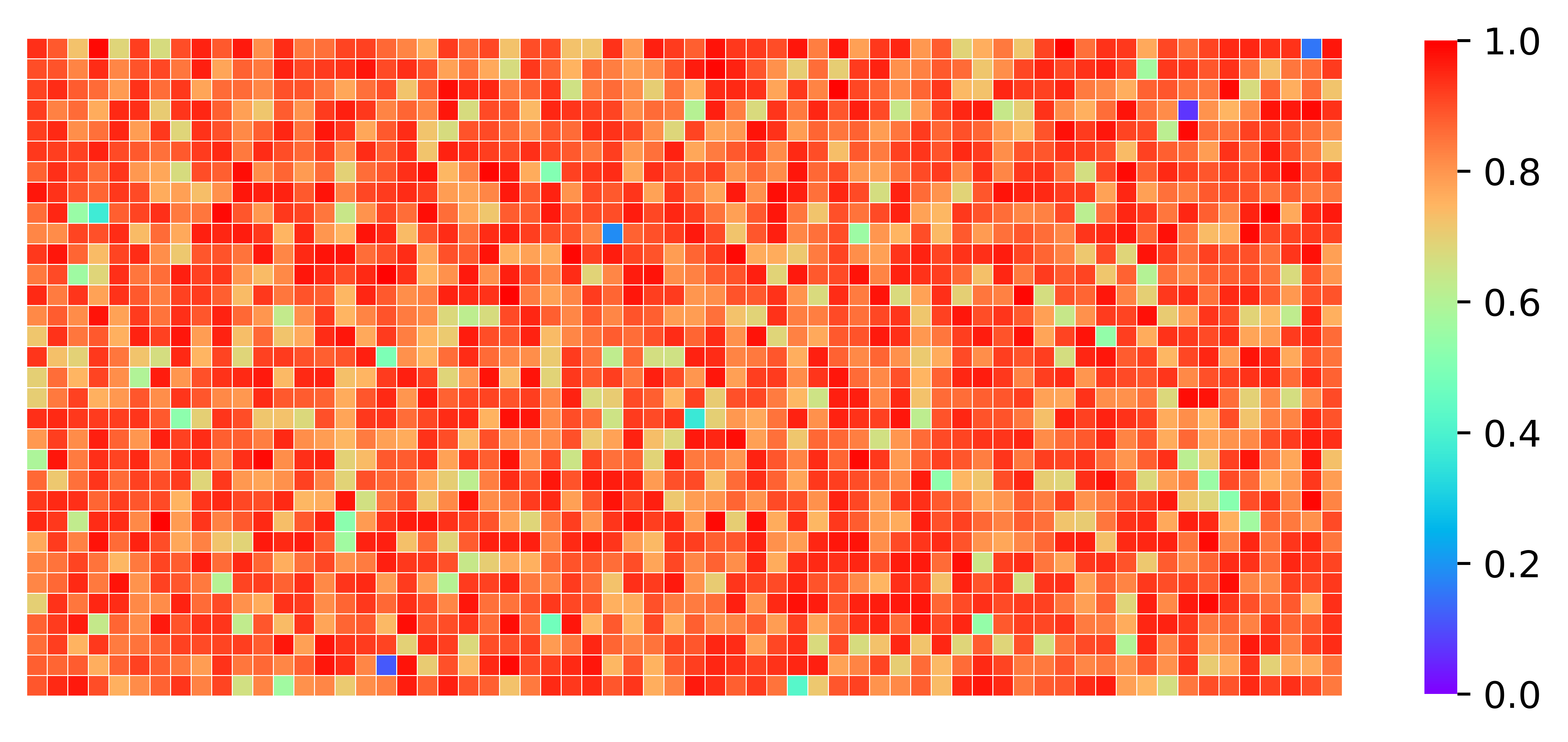}
    }\\
    \subfigure[$(B,H,W)$]{
        \includegraphics[width=0.45\linewidth]{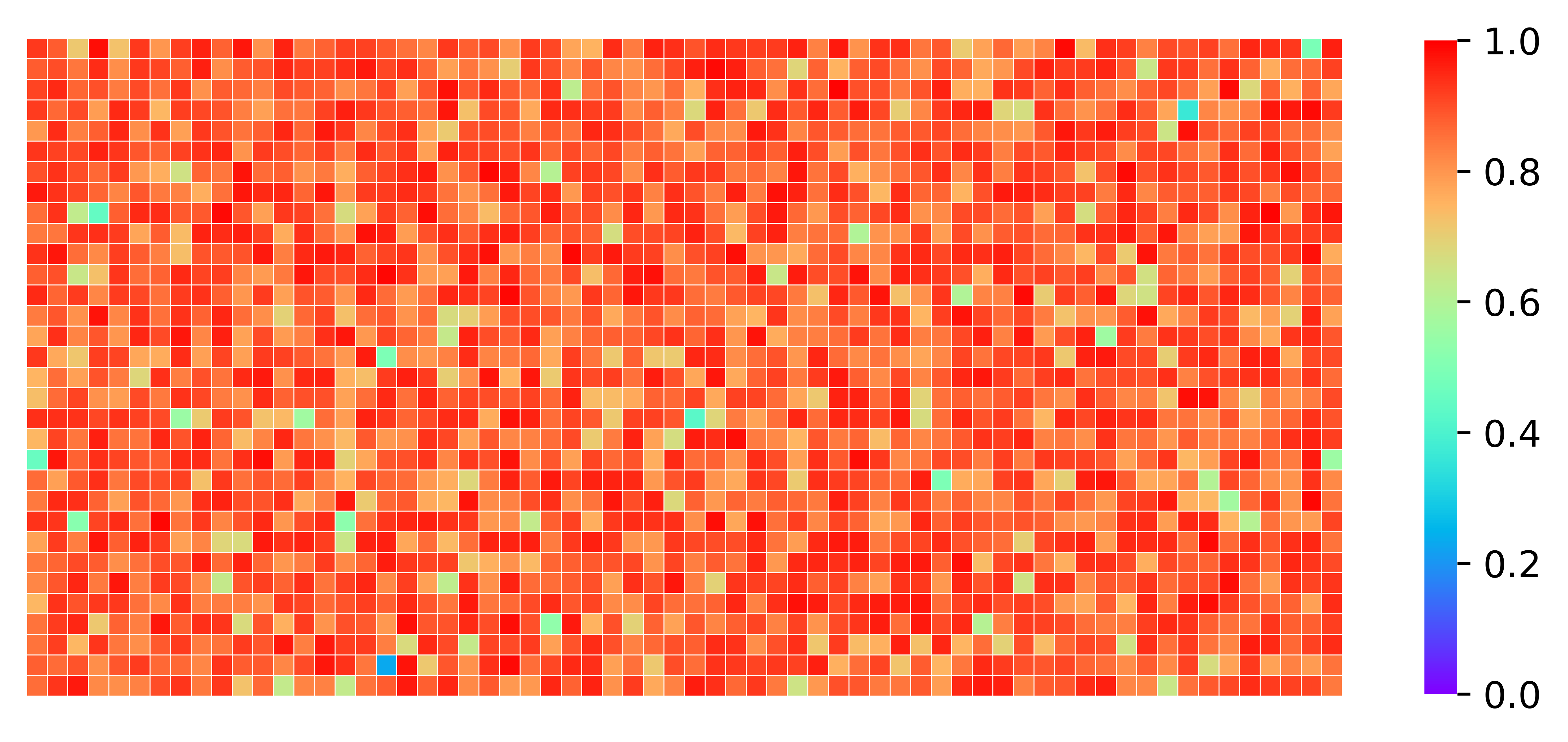}
    }
    \subfigure[$(H,W)$]{
        \includegraphics[width=0.45\linewidth]{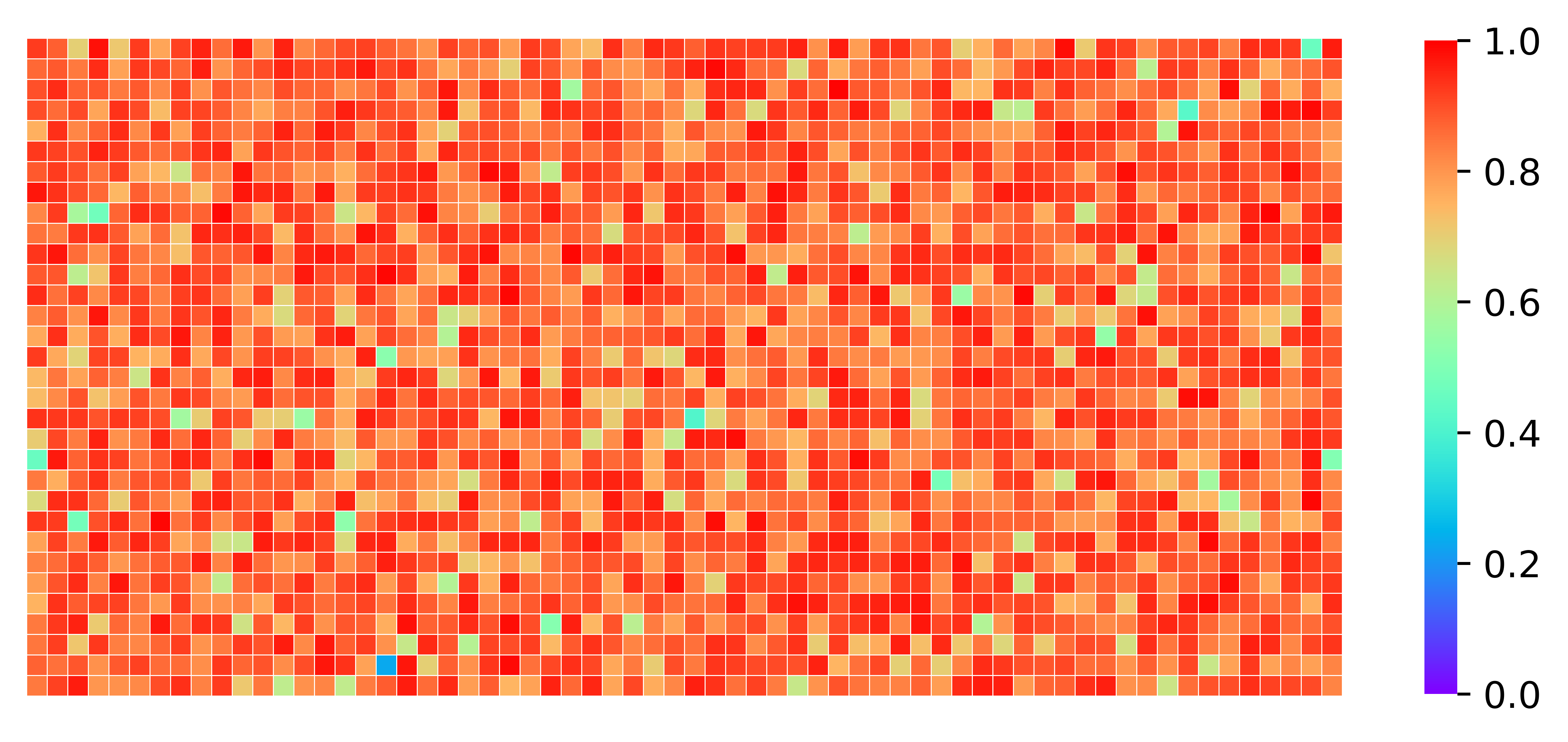}
    }
\caption{CKA similarity between the features of the student and the teacher.  We reshaped the CKA similarities of 2048 channels into a matrix of size $32 \times 64$. (a) shows distillation without normalization. (b), (c) and (d) show our NFD with normalization along different dimensions}
\label{fig:cka_heatmap}
\end{figure}

To further investigate the effect of NFD on feature distillation, we employ CKA \cite{CKA} to measure the similarity between the teacher and student features from the last layer of backbone network. We randomly sample 100 images from the Cityscapes validation set to calculate the CKA similarity. There is dimension alignment for student feature during distillation, so both the teacher and student features in this layer have 2048 channels. We compute the similarity for each channel, obtaining 2048 CKA similarities. 

For the qualitative analysis, we reshaped the CKA similarities of 2048 channels into a matrix of size $32 \times 64$, for better visualization in the the form of a heat map as shown in Fig.~\ref{fig:cka_heatmap}. It clearly shows that the proposed NFD always leads to a higher similarity between the features of the student and the teacher, no matter along which dimensions the normalization is performed.

\begin{figure}
    \centering
    \includegraphics[width=0.5\linewidth]{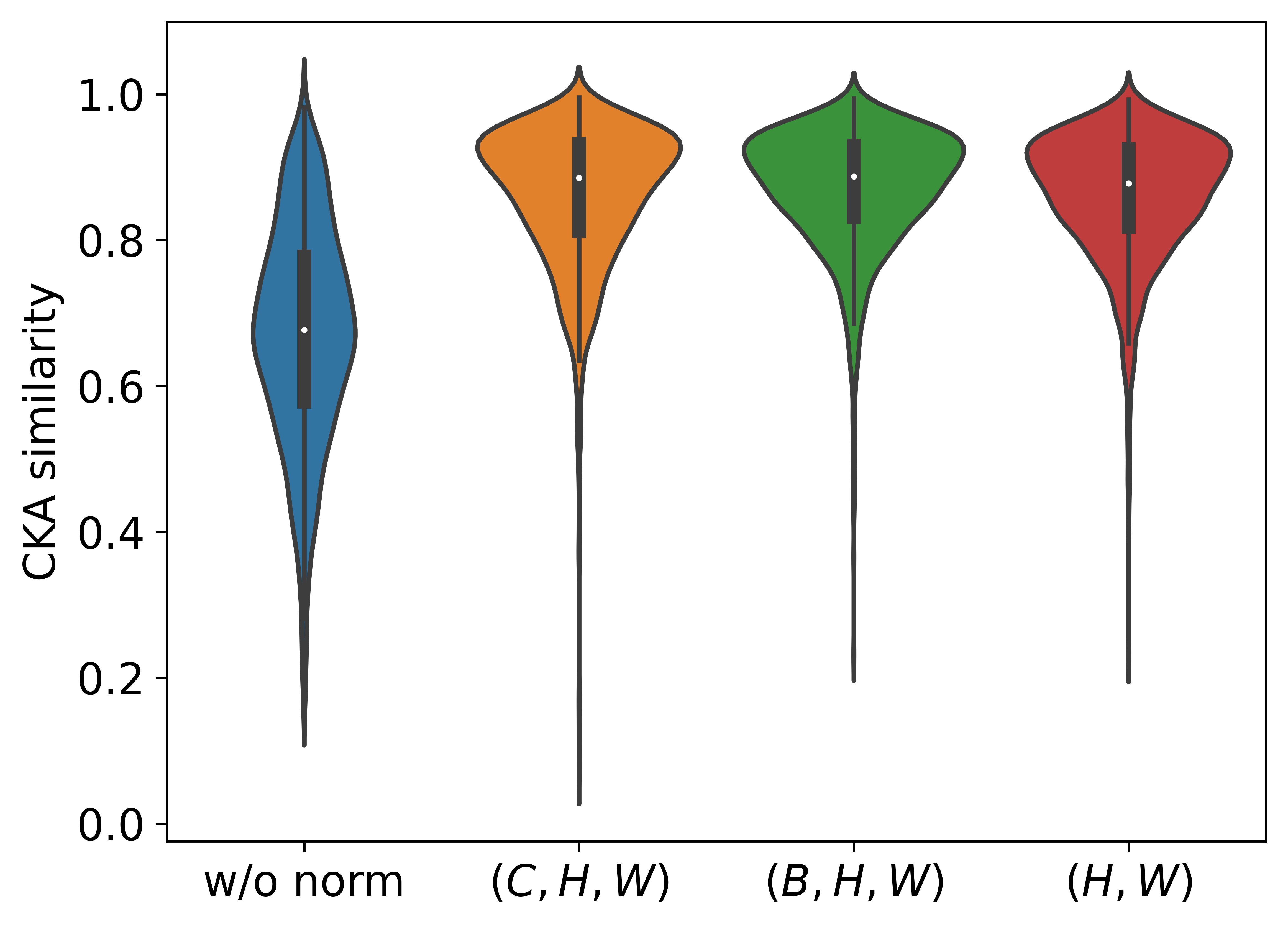}
\caption{The distribution of each channel's CKA similarity between the features of the teacher and student. ``w/o norm'' means without normalization, while ``$(C,H,W)$'', ``$(B,H,W)$'' and ``$(H,W)$'' denote normalization along different dimensions}
\label{fig:cka_violin}
\end{figure}

For the quantitative analysis, we analyze the distribution of CKA similarities of 2048 channels as shown in Fig.~\ref{fig:cka_violin}. For the proposed NFD, the CKA similarity of most channels falls in the range of $[0.8,1.0]$. In contrast, for the naive feature distillation, the CKA similarity is mainly distributed in the range of $[0.6,0.8]$. In addition, we calculate the average CKA similarity of all channels in Table~\ref{tab:average_cka}. The results show that the proposed NFD enables the student to better imitate the teacher, resulting in a higher feature similarity. 

\setlength{\tabcolsep}{4pt}
\begin{table}
    \begin{center}
    \caption{The average CKA similarity of all channels between the teacher and student features. ``w/o norm'' means NFD without normalization}
    \label{tab:average_cka}
    \begin{tabular}{cc}
        \hline
        Norm dimensions & Average CKA\\
        \hline
        w/o norm & 0.6711\\
        $(C,H,W)$ & 0.8612\\
        $(B,H,W)$ & 0.8705\\
        $(H,W)$ & 0.8611\\
        \hline
    \end{tabular}
    \end{center}
\end{table}
\setlength{\tabcolsep}{1.4pt}

\setlength{\tabcolsep}{4pt}
\begin{table}
    \begin{center}
    \caption{Ablation study on the validation set of Cityscapes about the distillation position. ``T'' denotes the teacher model, and ``S'' denotes the student model. ``backbone'' means the last layer of the backbone, ``decoder'' means the last layer of the decoder and ``logits'' means the final prediction layer. The performance of the teacher and the baseline student is shown in the
upper part of the table}
    \label{tab:ablation_feature_maps}
    \begin{tabular}{l|c|c}
        \hline
        Model & KD Position & mIoU (\%)\\
        \hline
        T: PSPNet-R101 & -- & 79.76\\
        S: PSPNet-R18 & -- & 72.65\\
        \hline
        \multirow{3}{*}{S: PSPNet-R18} & backbone & \textbf{76.63}\\
        & decoder & 75.58\\
        & logits & 75.62\\
        \hline
    \end{tabular}
    \end{center}
\end{table}
\setlength{\tabcolsep}{1.4pt}

\subsubsection{Position of Feature Distillation.} During feature distillation, the features can come from any available layers of the student and teacher networks. Here we choose three commonly used distillation positions: 1) the last layer of the backbone, 2) the last layer of the decoder (e.g., PPM \cite{PSPNet}), and 3) the final prediction layer. We use the features from the last layer of the backbone by default in the previous experiments, and now we present the results of NFD at different distillation positions in Table~\ref{tab:ablation_feature_maps}. Our method works well at different positions, especially at the last layer of the backbone. The features produced by the decoder or prediction layer are highly compressed and task-specific as they are located at the top level of the model. Instead, the features from the backbone tend to be more informative and rich in generic representations, which may explain the better results of NFD at the backbone. This result implies that our method has great potential for knowledge distillation in other vision tasks.

\begin{figure}
    \centering
    \includegraphics[width=0.6\linewidth]{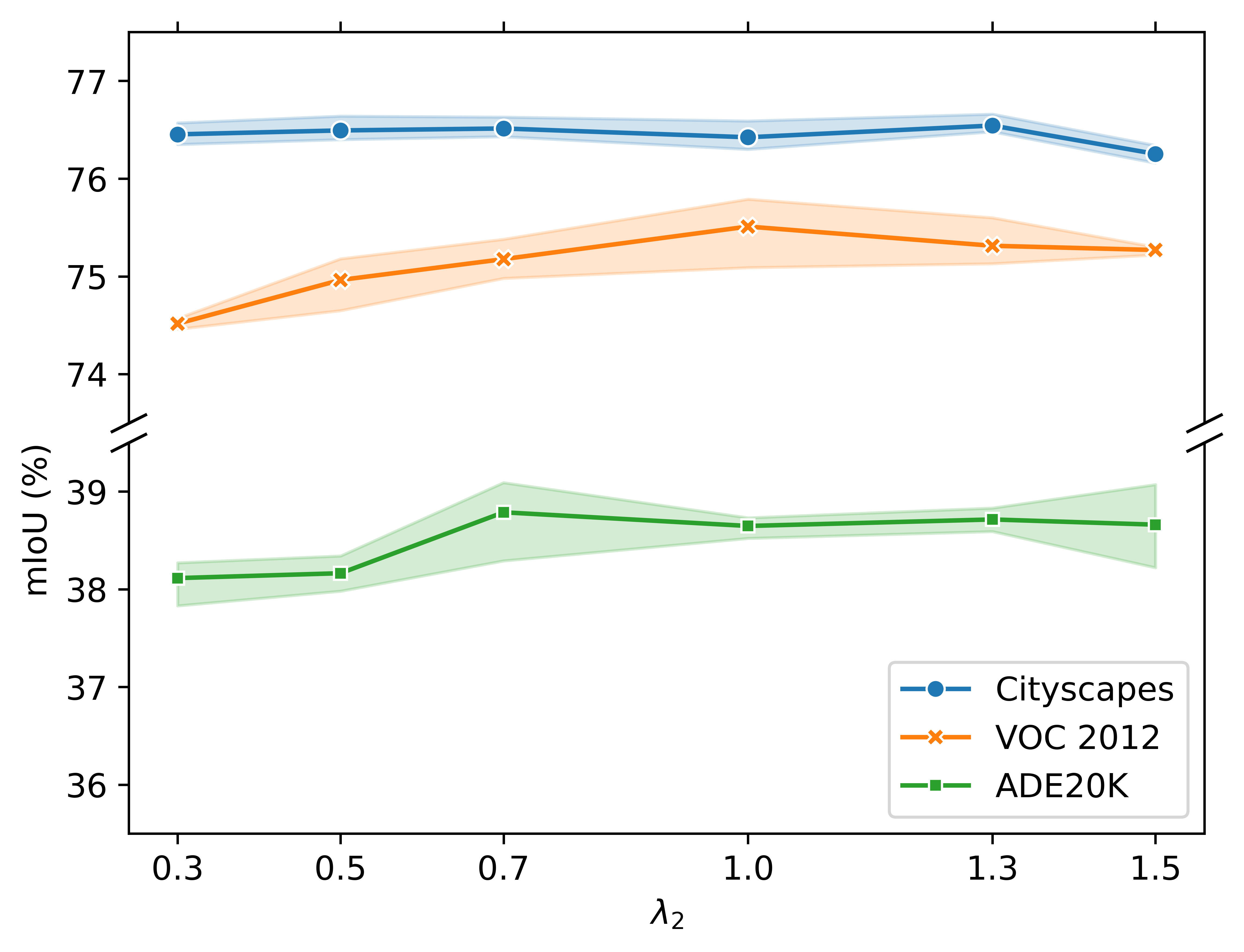}
\caption{Ablation study about the impact of $\lambda_2$ on the performance of our NFD. We run each experiment three times on Cityscapes, VOC 2012 and ADE20K datasets}
\label{fig:loss_weight}
\end{figure}

\subsubsection{Loss Weights.} The feature distillation loss in our method is weighted by the $\lambda_2$ in Eq.~(\ref{eq:total_loss}). Extensive experiments are conducted on Cityscapes, VOC 2012 and ADE20K datasets to investigate the sensitivity of the proposed NFD to the $\lambda_2$. We run each experiment three times, with $\lambda_2 \in \{0.3, 0.5, 0.7, 1.0, 1.3, 1.5\}$. The results in Fig.~\ref{fig:loss_weight} demonstrate the excellent robustness of the proposed NFD to hyper-parameters.

\subsubsection{Different Forms of Knowledge.} Previous method have designed various forms of knowledge extracted from the features, such as pair-wise similarity \cite{SKD}, intra-class feature variation \cite{IFVD} and channel-wise probability \cite{CWD}. Essentially, our method does not design a new form of knowledge, but rather removes the magnitude information from the original features, which we believe is not necessary for distillation. For a fair comparison of the effect of different forms of knowledge on feature distillation, we remove from all methods the parts that are irrelevant to feature distillation to conduct the experiments. Specifically, we remove the conventional KD loss \cite{KD} in \cite{SKD,IFVD,CWD} and the proposed NFD, and remove the adversarial distillation loss imposed on logits in \cite{SKD,IFVD,CWD}. The results are shown in Table~\ref{tab:ablation_knowledge}, where all methods use the feature from the last layer of the backbone network for distillation. In this experimental setup, the superiority of our method is more evident, demonstrating its simplicity and effectiveness. 

\setlength{\tabcolsep}{4pt}
\begin{table}
    \begin{center}
    \caption{Ablation study on the validation set of Cityscapes about different forms of knowledge. ``T'' denotes the teacher model, and ``S'' denotes the student model. The performance of the teacher and the baseline student is shown in the upper part of the table}
    \label{tab:ablation_knowledge}
    \begin{tabular}{l|l|c|c}
        \hline
        \multirow{2}{*}{Model} & \multirow{2}{*}{Knowledge} & \multicolumn{2}{c}{mIoU (\%)}\\
        \cline{3-4}
        & & Cityscapes & ADE20K\\
        \hline
        T: PSPNet-R101 & -- & 79.76 & 44.39\\
        S: PSPNet-R18 & -- & 72.65 & 35.03\\
        \hline
        \multirow{4}{*}{S: PSPNet-R18} & Pair-wise Similarity \cite{SKD} & 73.55 & 35.18\\
        & Intra-class Feature Variation \cite{IFVD} & 73.90 & 36.57\\
        & Channel-wise Probability \cite{CWD} & 73.98 & 35.91\\
        & Normalized Feature (Ours) & \textbf{75.86} & \textbf{38.55}\\
        \hline
    \end{tabular}
    \end{center}
\end{table}
\setlength{\tabcolsep}{1.4pt}

\section{Conclusion}
In this paper, we propose a simple yet effective feature distillation method for semantic segmentation called normalized feature distillation (NFD). Different from previous methods which strive to manually design various forms of knowledge, we focus on how to make distillation with original features effective without the need to design new forms of knowledge. We believe that forcing the student to imitate the exact same feature responses as the teacher is inefficient and unnecessary for distillation, but instead increases the learning difficulty. Therefore, the proposed NFD aims to remove the magnitude information from the original features by normalization during distillation. Extensive experiments for semantic segmentation demonstrate that our NFD can significantly improves the performance of the baseline student model. For future work, we will apply the proposed method to other vision tasks, such as image classification, object detection, and instance segmentation. Moreover, we consider applying our method to transformer networks.

% \clearpage
% ---- Bibliography ----
%
% BibTeX users should specify bibliography style 'splncs04'.
% References will then be sorted and formatted in the correct style.
%
\bibliographystyle{splncs04}
\bibliography{egbib}
\end{document}